\documentclass[letterpaper]{article} 
\usepackage{aaai25}  
\usepackage{times}  
\usepackage{helvet}  
\usepackage{courier}  
\usepackage[hyphens]{url}  
\usepackage{graphicx} 
\usepackage{natbib}  
\usepackage{caption} 
\usepackage{algorithm}
\usepackage{algorithmic}

\usepackage{graphicx}
\usepackage{booktabs}
\usepackage{amsmath}
\usepackage{amssymb}
\usepackage{multirow}
\usepackage{subcaption}
\usepackage{pifont}
\usepackage{colortbl}
\usepackage[table,xcdraw]{xcolor}
\usepackage{color}

\usepackage{esint}

\usepackage{newfloat}
\usepackage{listings}
\DeclareCaptionStyle{ruled}{labelfont=normalfont,labelsep=colon,strut=off} 
\floatstyle{ruled}
\newfloat{listing}{tb}{lst}{}
\floatname{listing}{Listing}
\title{ContextHOI: Spatial Context Learning for Human-Object Interaction Detection}
\author {
    Mingda Jia\textsuperscript{\rm 1},
    Liming Zhao\textsuperscript{\rm 2},
    Ge Li\textsuperscript{\rm 1}\thanks{Corresponding author},
    Yun Zheng\textsuperscript{\rm 2},
}
\affiliations {
    \textsuperscript{\rm 1}Peking University Shenzhen Graduate School\\
    \textsuperscript{\rm 2}Alibaba Group\\
    2201212832@stu.pku.edu.cn, geli@ece.pku.edu.cn, 
    \{lingchen.zlm, zhengyun.zy\}@alibaba-inc.com
}

\begin{document}

\maketitle

\begin{abstract}
Spatial contexts, such as the backgrounds and surroundings, are considered critical in Human-Object Interaction (HOI) recognition, especially when the instance-centric foreground is blurred or occluded. Recent advancements in HOI detectors are usually built upon detection transformer pipelines. While such an object-detection-oriented paradigm shows promise in localizing objects, its exploration of spatial context is often insufficient for accurately recognizing human actions. To enhance the capabilities of object detectors for HOI detection, we present a dual-branch framework named ContextHOI, which efficiently captures both object detection features and spatial contexts. In the context branch, we train the model to extract informative spatial context without requiring additional hand-craft background labels. Furthermore, we introduce context-aware spatial and semantic supervision to the context branch to filter out irrelevant noise and capture informative contexts. ContextHOI achieves state-of-the-art performance on the HICO-DET and v-coco benchmarks. For further validation, we construct a novel benchmark, HICO-$ambiguous$, which is a subset of HICO-DET that contains images with occluded or impaired instance cues. Extensive experiments across all benchmarks, complemented by visualizations, underscore the enhancements provided by ContextHOI, especially in recognizing interactions involving occluded or blurred instances.
\end{abstract}

%

\section{Introduction}
\label{sec:intro}
Human-object interaction (HOI) detection~\cite{gao2018ican} involves identifying instance locations and their interactions, typically represented as $<$Human, Interaction, Object$>$ triplets. Visual uncertainties in real-world scenes, such as occlusion, significantly affect the discernibility of the subjects. Thus, a primary challenge of HOI detection is accurately inferring the interactions with limited visual cues.

\begin{figure}[t]
  \centering
   \includegraphics[width=1.0\linewidth]{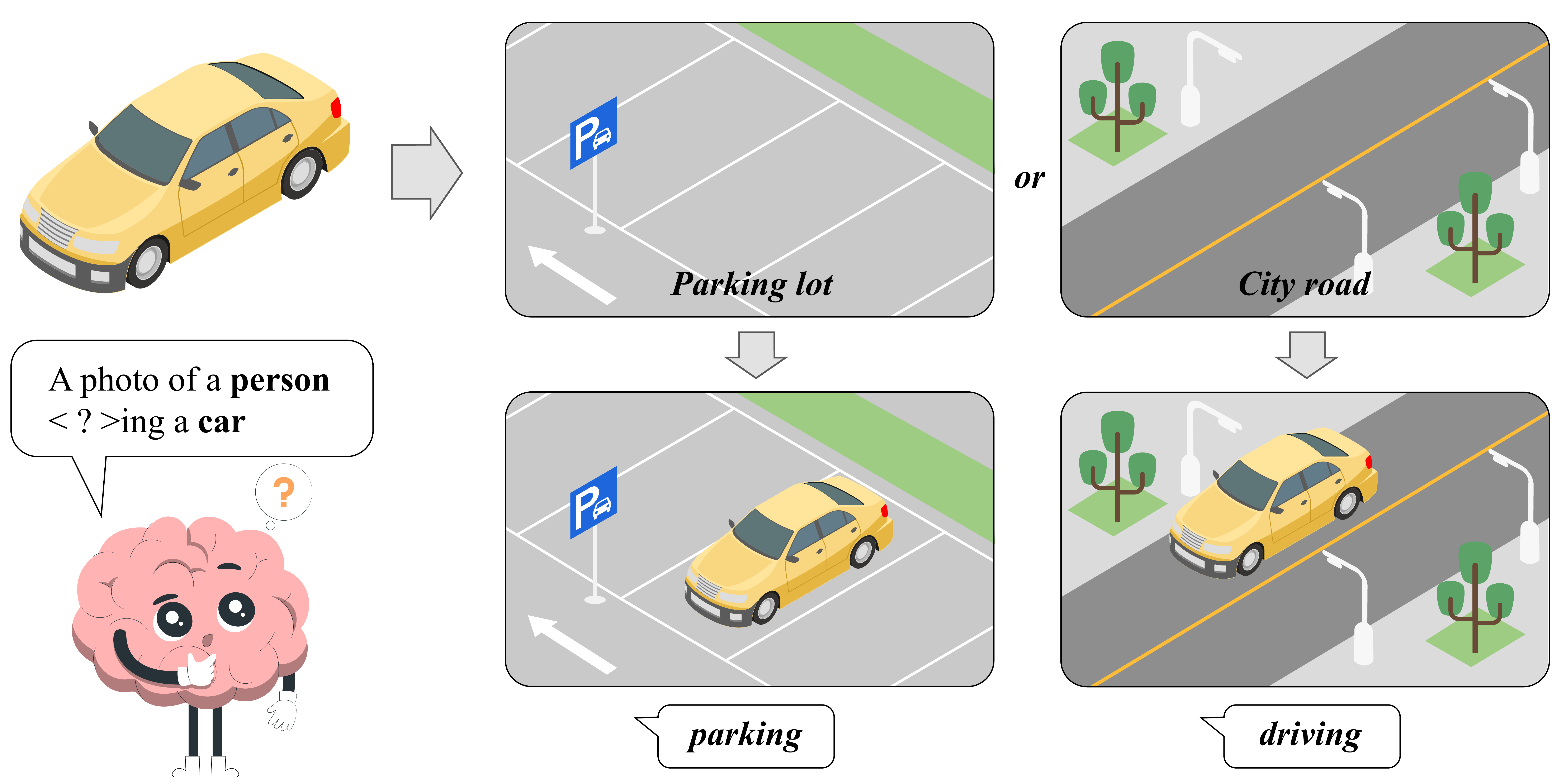}
   \caption{\textbf{The role of context learning in HOI Detection.} Spatial context, like a $parking lot$ or a $city road$ helps little with identify the salient car. However, context is critical in distinguishing human interactions. Both $parking$ and $driving$ are highly related to the context information. }
   \label{fig:figure_intro}
\end{figure} 

Recent advancements in HOI detection are driven by various approaches, including those based on convolutional networks~\cite{Wang_2019_ICCV}, and graphs~\cite{Gao-ECCV-DRG}. HOI detectors built upon transformers generally demonstrate superior performance~\cite{ning2023hoiclip}. Both one-stage and two-stage HOI detectors perform inference depending on the captured instance-centric attributes, while the two-stage methods rely more heavily on the confidence of their pre-trained object detection backbones~\cite{zhang2022upt}.

Despite significant advancements, existing HOI detectors face a major difficulty. They are vulnerable in \textbf{identifying HOI scenes with limited foreground visual cues}, such as when subjects are blurred or occluded. In contrast, Humans can accurately recognize HOIs even when instances are unclear or completely unseen. For example, we can easily infer the presence of a driver driving in a car on the highway, though the driver is obscured by the tinted windows of the car, as illustrated in Figure~\ref{fig:figure_intro}.


Upon analysing such difficulties, we observe a potential tendency to \textbf{overemphasize instance-centric attributes} in previous HOI detectors. 
 Detection transformers (DETR), foundational to many HOI detectors, often capture minimal context information while neglecting most backgrounds as negative or irrelevant. Although this approach is a consensus of $object$ $detection$, it may not be sufficient enough for robust $action$ $recognition$, which requires a more comprehensive scene understanding. Therefore, the core issue appears to be the misalignment between the minimal context modelling inherent in object detection and the extensive contextual requirements essential for HOI detection. While acknowledging the impressive capabilities of transformer-based object detectors, our goal is to adapt them to HOI detection better.

A practical adaption strategy is augmenting instance features from object detectors with additional contextual information. Hence, we propose $\textrm{ContextHOI}$, a dual-branch framework that integrates object detection features with learned context to enhance the robustness of HOI predictions. Specifically, we design multiple novel modules in the context learning branch to efficiently capture informative contextual features.


In the context branch, we propose a context extractor that grounds context features within the image to capture the context regions. Recent HOI detectors have started to capture global visual features and learn backgrounds heuristically. However, these approaches might degrade into mere replications of instance-centric modules due to the lack of context-aware supervision. To mitigate such risk, we incorporate explicit spatial and semantic supervision into the context extractor. These supervisions enable efficient context learning without additional hand-craft context labels or segmentation priors.

For the \textbf{spatial supervision}, we propose a series of self-supervised $spatially$ $contrastive$ $constraints$. These constraints provide multi-level coarse-to-fine constraints, aiming to increase the margin between context and instance regions. Such an objective guides the context branch to focus on regions out of RoIs while minimising the background noise for the instance branch. Furthermore, we propose a dynamic distance weight in one of the constraints, which allows the context extractor to capture contexts with suitable spatial distributions rather than regressing regions near the image margins. 

For the \textbf{semantic supervision}, we introduce a $semantic$-$guided$ $context$ $explorer$. Fascinated by the significant capability of pre-trained VLMs~\cite{clip, EVA-CLIP} in representing the context~\cite{an2023context}, we integrate their prior knowledge into our context explorer. Additionally, we introduce an adaptive knowledge distillation approach to bridge the gap between the embedding space of VLMs and the feature space of our visual encoder.



We introduce a context aggregator to extract further more crucial context features that can complete the detection features. The context aggregator grounds the sampled context features according to the object detection features while enhancing the communication between the local and global information. Finally, these features are fused for the interaction prediction.

Moreover, we propose a new benchmark, HICO-DET (ambiguous), for evaluating HOI detectors on scenes with unclear foreground visual contents. ContextHOI demonstrates competitive performance on the regular setting of HICO-DET and v-coco while demonstrating significantly stronger robustness than previous HOI detectors on HICO-DET (ambiguous). Additionally, ContextHOI also obtains zero-shot ability.
 
\noindent{In this paper, our contributions are threefold:}
\begin{itemize}
\setlength{\itemsep}{0pt}
\setlength{\parsep}{0pt}
\setlength{\parskip}{0pt}
\item We revisit and explore the significance of spatial contexts in HOI detection, which is not sufficiently discussed in recent HOI methods.
\item To the best of our knowledge, we are the first to systematically learn spatial contexts for enhancing human-object interaction detection. Our model can be trained to capture informative context features without additional background labels or segmentation priors. 
\item ContextHOI achieves competitive results on HICO-DET and v-coco while obtaining significant state-of-the-art performance on HICO-DET (ambiguous), containing HOI scenes with unclear or occluded foregrounds, highlighting the robustness of our approach.
\end{itemize} 

 \begin{figure*}[t]
  \centering
   \includegraphics[width=1.0\linewidth]{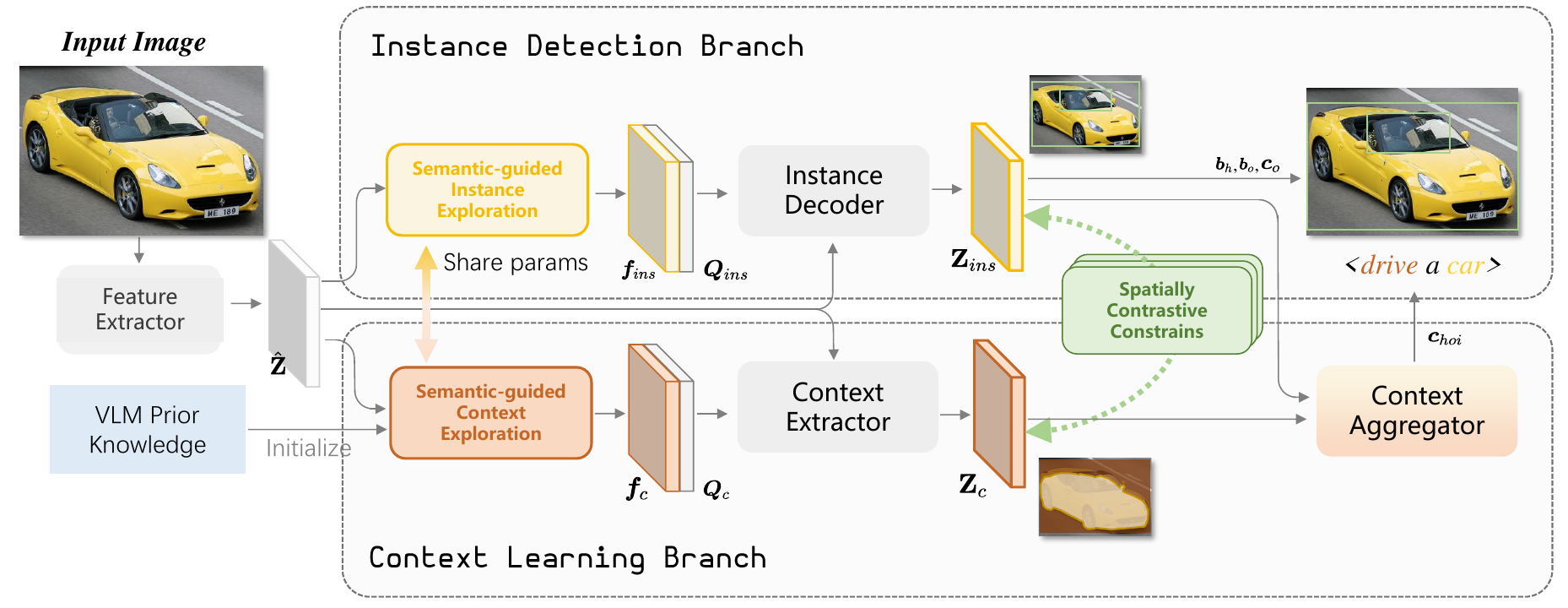}
   \caption{\textbf{The overall architecture of ContextHOI.} ContextHOI has a dual-branch and fusion structure, with instance detection and context learning branches. The instance detection branch captures instance-centric attributes, while the context learning branch focuses on instance-independent context features. We introduce a semantic-guided instance/context exploration module to distil prior knowledge from VLM to help ground informative visual content. A set of spatially contrastive constraints supervises the learned instances and contexts to focus on different visual aspects. Finally, a context aggregator will fuse the instance and context feature for HOI prediction. }
   \label{fig:network}
\end{figure*} 
\section{Related Works}
\label{sec:related works}

\textbf{HOI detection.} Based on transformers, one-stage HOI detectors \cite{tamura_cvpr2021,ning2023hoiclip} predict HOIs with shared~\cite{kim2021hotr}, parallel~\cite{ kim2023relationalMuren}, or sequential~\cite{NEURIPS2021_8f1d4362} transformers. In contrast, two-stage detectors~\cite{zhang2022upt, ada_cm} utilize a pre-trained object detector backbone to capture instance locations and foreground attributes at the first stage. In the second stage, they predict interactions using specialized modules. 

However, both one-stage and two-stage HOI methods depend heavily on transformer-based object detection pipelines, which limit their ability to model spatial contexts. Thus, these HOI detectors often struggle to recognize images with unclear foregrounds. This limitation is more critical in two-stage approaches, as detailed in Table~\ref{tab:performance_hico_ambiguous}. Such issues are foreseen by CDN~\cite{NEURIPS2021_8f1d4362}. However, the exploration of systematically integrating context learning into HOI remains unexplored.

\noindent\textbf{Spatial context learning in HOI.} Spatial contexts are crucial in relative tasks like object detection~\cite{Chen_2018_ECCV}, scene-graph generation~\cite{yang2018graph, zhai2023soarscenedebiasingopensetaction} and group recognition~\cite{Yuan_Ni_2021}. 

In HOI detection, several works capture contexts with customized spatial modules. Early methods used dense graphs~\cite{Gao-ECCV-DRG,zhang2021scg} for global context. Later approaches focused on instance-centric contexts~\cite{gao2018ican, Wang_2019_ICCV}. Recently, BCOM~\cite{bcom_Wang_2024_CVPR} attempted to enhance instance features through occluded part extrapolation (OPE). However, they still depended heavily on object detection features. In contrast, our method extracts context features independently of instances. Such an approach ensures the focused context regions are not limited to instance-centric RoIs, providing a broader understanding of the scene. Secondly, it mitigates the error accumulation from failures in object detection backbones.

As for the semantic-guided learning approaches, some work~\cite{yuan2022rlip, Yuan_2023_ICCV} leverages utilize learnable text embeddings to represent pseudo background categories, while the others~\cite{ada_cm,bcom_Wang_2024_CVPR} learn a global concept memory to explore background information. However, their approaches are implicit or heuristic. The lack of explicit spatial constraints might downgrade their context-aware modules to mere repetitions of instance-centric modules. To address these limitations, we propose explicit spatial supervision as context-centric guidance.



\section{Method}
This section details ContextHOI, a framework that learns informative spatial context for HOI detection. In section 3.1, we outline the overall architecture. In sections 3.2 and 3.3, we deliver the design of the two supervisions on the context extractor, the spatially contrastive constraints, and the semantic-guided context explorer. In section 3.4, we present the context aggregator. Finally, we introduce the training method in section 3.5

\subsection{Overall Architecture.}
\label{sec:3.1}
The network design of ContextHOI is illustrated in Figure~\ref{fig:network}, our framework follows a parallel dual-branch architecture, with an instance detection branch and a context learning branch. Given an input image, we first utilize a pre-trained visual encoder~\cite{detr} to summarize the image feature map $\hat{\mathbf{Z}}$. The image feature will be fed into both branches as the visual content memory. 

In the instance detection branch, an instance decoder with several transformer decoder layers will utilize a set of instance queries $\boldsymbol{Q}_{ins}\in\mathbb{R}^{2N_q\times{C}}$ to ground $\hat{\mathbf{Z}}$ and generate instance-centric features ${\mathbf{Z}_{ins}}$. Several prediction heads take the captured feature and generate predictions for detection-oriented tasks, including the human bounding box $\boldsymbol{B}_{h}\in\mathbb{R}^{N_q\times{4}}$, object bounding box $\boldsymbol{B}_{o}\in\mathbb{R}^{N_q\times{4}}$ and object categories $\boldsymbol{C}_{o}\in\mathbb{R}^{N_q\times{N_o}}$, where $N_q$ is the number of queries, $N_o$ is the number of object classes. 

In the context learning branch, our context extractor, which shares the same architecture as the instance decoder, captures context feature ${\mathbf{Z}_{c}}$ by the context queries $\boldsymbol{Q}_{c}\in\mathbb{R}^{N_q\times{C}}$. Then, our context aggregator integrates the instance-centric feature and contexts. A single prediction head will take the aggregated feature and predict the hoi categories $\boldsymbol{C}_{hoi}\in\mathbb{R}^{N_q\times{N_{hoi}}}$, where $N_{hoi}$ are the number of HOI triplet combinations.

\subsection{Spatially Contrastive Constraints.}
\label{sec:3.2}
In the following part, we propose a set of spatially contrastive constraints in three coarse-to-fine supervision levels, including feature-level constraint $\mathcal{L}_{FC}$, region-level constraint $\mathcal{L}_{RC}$ and instance-level constraint $\mathcal{L}_{IC}$. These constraints work as training loss on the outputs of the instance decoder and context extractor simultaneously. The main objective of these constraints is to diverge the attention areas of the instance decoder and context extractor. Furthermore, we introduce dynamic distance weight into $\mathcal{L}_{IC}$ to maintain the suitable spatial shape of the learned contexts.

\noindent\textbf{Feature-level constraint.} Given the instance features $\mathbf{Z}_{ins}$ and the context features $\mathbf{Z}_{c}$, we calculate the absolute value of their mean cosine similarity along the query dimension. The feature patch with much higher similarity to the corresponding instance patches is seen as a propagation negative through the context features. $\mathcal{L}_{FC}$ can be expressed by:
\begin{equation}
    \begin{aligned}
           \mathcal{L}_{FC} = 
          \dfrac{1}{\left |\bar{\boldsymbol{\Phi}}\right | }
          \sum_{k=1}^{N_q}   
        \dfrac{\left |{{\hat{\boldsymbol{z}}^{k\top}_{ins}}\hat{\boldsymbol{z}}_{c}^k} \right |} {\|\hat{\boldsymbol{z}}^{k\top}_{ins}\|_1 \|\hat{\boldsymbol{z}}_{c}^k\|_1 + \epsilon },
        k \notin \boldsymbol{\Phi},
    \end{aligned}
    \label{eq:featureloss}
\end{equation}
in which the summarized instance and context features from dual-decoders are reshaped to patched feature tokens $\{ \hat{\boldsymbol{z}}_{ins}^k \}_{k = 1}^{N_q} \in \mathbb{R}^{L_{dec}\times{C}}$ and $\{ \hat{\boldsymbol{z}}_{c}^k\}_{k = 1}^{N_q} \in \mathbb{R}^{L_{dec}\times{C}}$. $L_{dec}$ is the number of decoder layers shared by instance decoder and context extractor. We do not calculate $\mathcal{L}_{FC}$ on $\boldsymbol\Phi$, which are features indicated by zero-padded queries. 
\begin{figure}[t]
  \centering
   \includegraphics[width=0.95\linewidth]{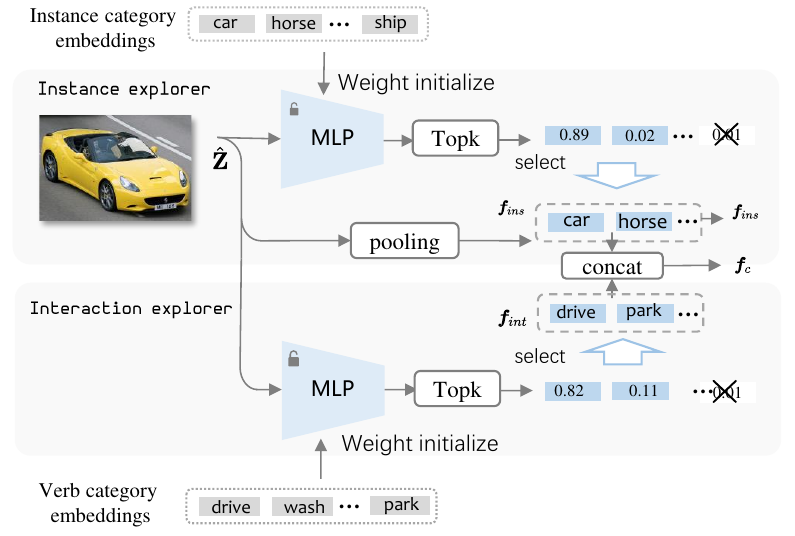}
   \caption{\textbf{Inner design of semantic-guided context exploration} module. $\textrm{pooling}$ refers to mean pooling on the spatial dimension of $\hat{\mathbf{Z}}$, $\textrm{concat}$ refers to concatenation.}
   \label{fig:hcr}
\end{figure} 

\noindent\textbf{Region-level constraint.}
At the region level, we constrain the learned positional guided embeddings $\boldsymbol{P}_{ins}\in \mathbb{R}^{N_q\times{C}}$ 
 and $\boldsymbol{P}_{c}\in \mathbb{R}^{N_q\times{C}}$ predicted together with instance and context features by corresponding decoders. As the guided embeddings take the role of directing transformer attention to specified local or global regions~\cite{liao2022gen}, we constrain $\boldsymbol{P}_{ins}$ and $\boldsymbol{P}_{c}$ to be distinct in the query position domain by reversed L1-distance. This constraint includes an $\exp$ operation to maintain a non-negative loss form. The constraint can be formulated as: 
\begin{equation}
    \begin{aligned}
          \mathcal{L}_{RC} = 
          \dfrac{1}{N_q }
          \sum_{k=1}^{N_q} 
 \exp({-\left \|{\boldsymbol{p}}_{ins}^k-\boldsymbol{p}_{c}^k  \right \| _1)},
    \end{aligned}
    \label{eq:regionloss}
\end{equation}
in which $\boldsymbol{p}_{ins}^k$ and $\boldsymbol{p}_{c}^k$ are instance-guided embedding and context-guided embedding at query position $k$. 

\noindent\textbf{Instance-level constraint.}
Aiming to capture $\mathbf{Z}_{c}$ with structured optimization goals, we extend the context extractor with a simple MLP, predicting context region boxes $\boldsymbol{B}_c \in \mathbb{R}^{N_q\times{4}}$. Then we reverse the Generalized Intersection-over-Union (GIoU) \cite{giou}  to formulate the loss, calculated on predicted $\hat{\boldsymbol{B}}_c$ and ground-truth $\boldsymbol{B}_h$ and $\boldsymbol{B}_o$. However, these constraints are too strict, which causes the context to shift to the edges of the images. To address this, we introduce a dynamic distance weight to improve context learning. This weight is calculated as:
\begin{equation}
 \mathcal{W}_{d}(\boldsymbol{b}_i, \boldsymbol{b}_j) = \exp(-\frac{\left | \boldsymbol{b}_i - \boldsymbol{b}_j\right|}{ \tau + \epsilon} ),
 \end{equation}
 \label{eq:distance_weight}
where $\tau$ is a learnable parameter describing a dynamic margin between the considered boxes. With this weight, the instance-level constraint can be expressed by:
\begin{equation}
    \begin{aligned}
           \mathcal{L}_{IC} = 
          \dfrac{1}{2\left |\bar{\boldsymbol{\Phi}}\right |}
          \sum_{k=1}^{N_q}
            \left[2 + \mathcal{W}_{d}(\boldsymbol{b}_h, {\hat{\boldsymbol{b}}}_c^{k})\textrm{GIoU}\left( \boldsymbol{b}_h, {\hat{\boldsymbol{b}}}_c^{k}\right)
              \right.\\
            \left. +\mathcal{W}_{d}(\boldsymbol{b}_o, {\hat{\boldsymbol{b}}}_c^{k})\textrm{GIoU}\left( \boldsymbol{b}_o, \hat{\boldsymbol{b}}_c^{k}\right)
            \right], k\notin\boldsymbol{\Phi},
    \end{aligned}
    \label{eq:instanceloss}
\end{equation}
 in which $\boldsymbol{b}_h$, $\boldsymbol{b}_o$ and $\hat{\boldsymbol{b}}_c^{k}$ are ground-truth human box, ground-truth object box and corresponding context coordinate predictions at query dimension $k$. Thanks to the dynamic distance weight, the strength of the positive signal from instance-level constraint diminishes significantly as the captured context regions approach the edges of the image. 

\subsection{Semantic-guided Context Exploration.}
\label{sec:3.3}
The semantic-guided context exploration module, detailed in Figure~\ref{fig:hcr}, provides semantic guidance to the following context extractor. Specifically, we construct learnable instance- and interaction-explorers in the exploration module. To distill VLM knowledge to these explorers, we initialize their weights using VLM text embeddings of object and verb categories. To mitigate the gap between the VLM text embedding space and the feature space of the visual extractors, we do not utilize explorers directly as classification heads. Instead, these explorers compute the VL similarity between visual features and linguistic representations. This process can be formulated as follows:
\begin{equation}
    \begin{aligned}
            {\omega}_{ins} &=  \sigma\textrm{MLP}_{vlm}^{ins}(\hat{\mathbf{Z}}, \theta_{ins}), \\
            {\omega}_{int} &=  \sigma\textrm{MLP}_{vlm}^{int}(\hat{\mathbf{Z}}, \theta_{int} ),
    \end{aligned}
    \label{eq:ccr_2}
\end{equation}
where $\theta_{ins}$ and $ \theta_{int}$ refer to the weights of the instance explorer and interaction explorer, while $\sigma$ refers to a Gumbel \textrm{softmax}~\cite{jang2017categoricalreparameterizationgumbelsoftmax}, which optimizes the tuning of explorers. ${\omega}_{ins}$ and ${\omega}_{int}$ refer to the category-related similarity of the visual content memory. Then, we select $N_q$ pooled feature maps with the highest pooled similarities $\hat{\omega}_{ins}$ and $\hat{\omega}_{int}$ in a Topk manner. The selected features are used as online instance-aware guidance and interaction-aware guidance $\boldsymbol{f}_{ins}\in\mathbb{R}^{N_q\times{C}}$ and $\boldsymbol{f}_{int}\in\mathbb{R}^{N_q\times{C}}$. Such process is expressed by:
\begin{equation}
    \begin{aligned}
            \boldsymbol{f}_{ins} &= \prod_{k = 0}^{N_q} \scriptsize{\begin{matrix} 
              \textrm{Topk}\\ \hat{\omega}_{ins}^k \in {N_{o}}
            \end{matrix}} {(\hat{\mathbf{Z}}_{{\omega}_{ins}}^k)}, \\
            \boldsymbol{f}_{int} &=  \prod_{k = 0}^{N_q}  \scriptsize{\begin{matrix} 
              \textrm{Topk}\\ \hat{\omega}_{int}^k \in {N_{v}}
            \end{matrix}}{(\hat{\mathbf{Z}}_{{\omega}_{c}}^k)},
    \end{aligned}
    \label{eq:ccr_3}
\end{equation}
where $\prod$ refers to concatenation along the query dimension. The query dimension is padded with zero if the number of predicted similarities $N_o$ or $N_c$ are less than $N_q$. To our best practice, we select OpenAI CLIP~\cite{clip} as our semantic teacher for the explorers. As both instance and interaction-related semantic information help capture meaningful contexts, we concatenate $\boldsymbol{f}_{ins}$ and $\boldsymbol{f}_{int}$ in the query dimension to construct semantic context guidance $\boldsymbol{f}_{c}$. Finally, we add $\boldsymbol{f}_{c}$ to context queries. Observing that such semantic guidance is also helpful for instance detection, we incorporate $\boldsymbol{f}_{ins}$ with the instance queries, this process can be seen as an additional semantic-guided instance exploration for instance detection branch, shown in Figure~\ref{fig:network}. In practice, we employ two linear layers to map the semantic guidance to align with the query dimensions of the instance and context queries.
\begin{figure}[t]
  \centering
   \includegraphics[width=1.0\linewidth]{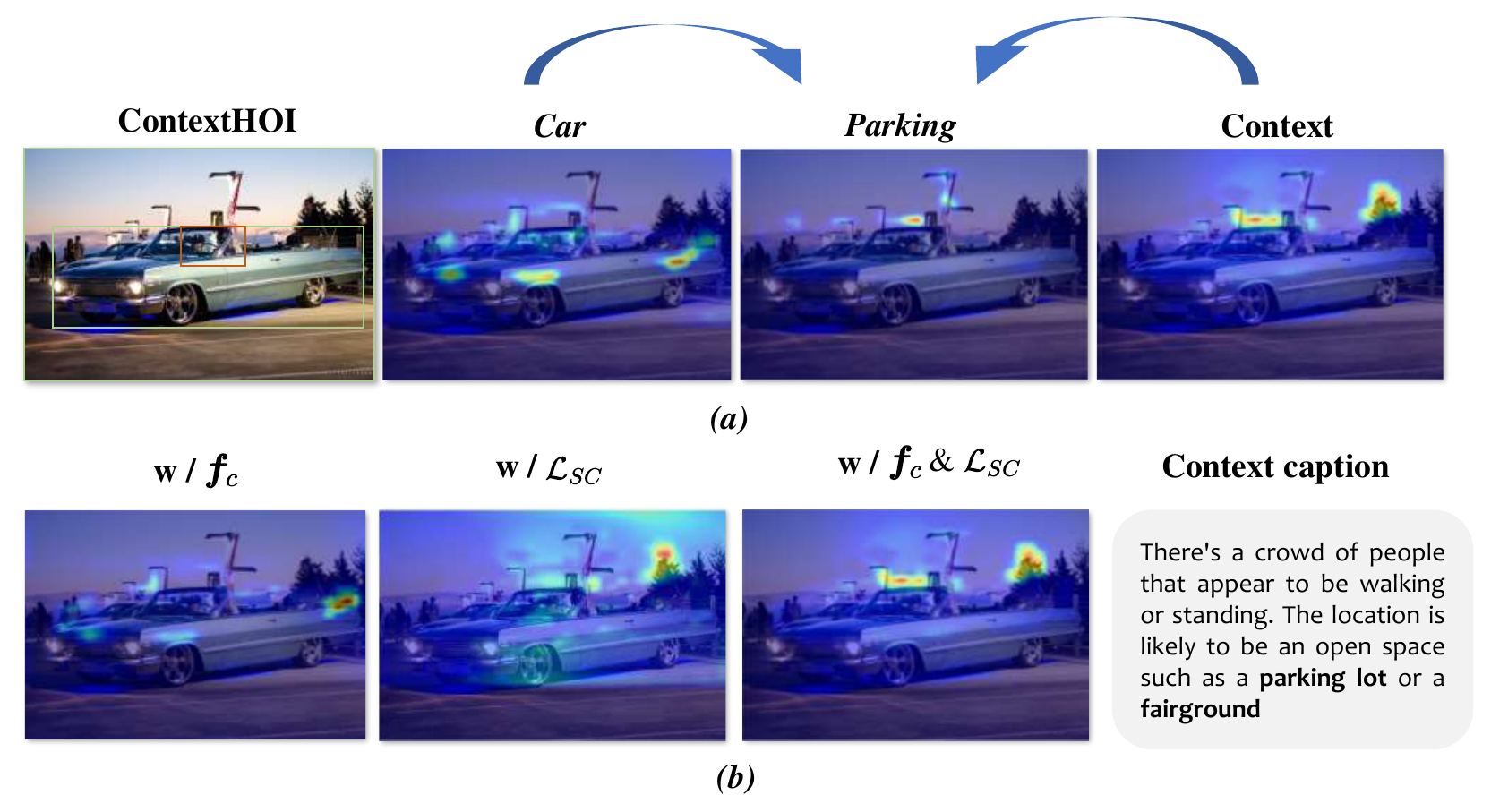}
   \caption{\textbf{Visualization analysis on spatial context learning.} (a) The feature map of the last layer of instance decoder, context aggregator and context extractor, indexed by the highest logits. Our instance decoder focuses on the appearance of the car, and the context extractor captures backgrounds and surrounding humans. (b) The features that are captured by context extractors with different component compositions. Both components help capture spatial contexts. Best viewed in color. Please zoom in for details.}
   \label{fig:ablation_heatmap}
\end{figure} 
\subsection{Context Aggregator}
\label{sec:3.4}
Given the captured instance detection feature ${\mathbf{Z}_{ins}}$ and informative contexts ${\mathbf{Z}_{c}}$, a learnable aggregation query $\boldsymbol{Q}_{agg}\in\mathbb{R}^{N_q\times{C}}$ is learned to fuse instance and context feature that complementary to each other. Inspired by the knowledge integration approach in ~\cite{ning2023hoiclip}, we implement the aggregator with a multi-branch transformer decoder with shared cross-attention weight. In each decoder layer, such process can be expressed by:
\begin{equation}
    \begin{aligned}
            \hat{\mathbf{Z}}_{ins} &= \textrm{CrossAttn}(\boldsymbol{Q}_{agg}, {\mathbf{Z}_{ins}}, \theta_{attn}), \\
            \hat{\mathbf{Z}}_{c} &= \textrm{CrossAttn}(\boldsymbol{Q}_{agg}, {\mathbf{Z}_{c}}, \theta_{attn}). \\
    \end{aligned}
    \label{eq:ccr_4}
\end{equation}
We also calculate a cross-attention between the VLM visual feature and the aggregation query as an auxiliary semantic guidance, obtaining $\hat{\mathbf{Z}}_{vlm}$. After this step, we concat $\hat{\mathbf{Z}}_{ins}$, $\hat{\mathbf{Z}}_{c}$ and $\hat{\mathbf{Z}}_{vlm}$ as the aggregated feature for interaction prediction.

\subsection{Training}
\label{sec:3.5}
\textbf{Context-aware training.} ContextHOI is trained end-to-end and predict HOIs with set prediction manner following~\cite{kim2021hotr}. During training, we calculate the matching cost between all the predictions and ground truths with a bipartite graph, under the Hungarian Matching algorithm~\cite{Chen2021ReformulatingHD}, and select matched predictions with the lowest matching costs. The matched predictions are combined to a set $\left \{\boldsymbol{B}_{h}, \boldsymbol{B}_{o}, \boldsymbol{C}_o, \boldsymbol{C}_{hoi}\right \}$, and the conventional HOI losses $\mathcal{L}_{HOI}$ proposed by QPIC~\cite{tamura_cvpr2021} are calculated on them. 

Additionally, for the spatially contrastive constraints proposed in section 3.2, the formulated losses are as follows: \\
\begin{equation}
    \begin{aligned}
         \mathcal{L}_{SC} =  \lambda_{fc}\mathcal{L}_{FC} + \lambda_{rc}\mathcal{L}_{RC}+ \lambda_{ic}\mathcal{L}_{IC}.
    \end{aligned}
    \label{eq:conetxtloss}
\end{equation}
 During the training stage, all the parameters share gradient propagation with the following end-to-end form: \\
\begin{equation}
    \begin{aligned}
          \mathcal{L} = \mathcal{L}_{HOI} + \mathcal{L}_{SC}.
    \end{aligned}
    \label{eq:totalloss}
\end{equation}

\begin{table}[H]
\centering
\scalebox{0.78}{
\begin{tabular}{cccccc}
\midrule
\multicolumn{1}{c}{\multirow{2}{*}{Method}} &
  \multicolumn{1}{c}{\multirow{2}{*}{Backbone}} &
  \multicolumn{3}{c}{HICO-DET (Default)} &
  \multicolumn{1}{c}{v-coco} \\
  \multicolumn{1}{c}{} &
\multicolumn{1}{c}{} &
 \multicolumn{1}{c}{\textit{full}} &
  \multicolumn{1}{c}{\textit{rare}} &
  \multicolumn{1}{c}{\textit{non-rare}} &
  \multicolumn{1}{c}{\textit{$AP_{role}^{\#1}$}}\\ \midrule
\specialrule{0em}{1.5pt}{1.5pt}
\midrule
\multicolumn{1}{c}{iCAN \citep{gao2018ican}} &
  \multicolumn{1}{c}{R50} & 
   14.84 &
  10.45 & 
  16.15 & 
  45.3 \\ 
  \multicolumn{1}{c}{\cite{Wang_2019_ICCV}} &
  \multicolumn{1}{c}{R50} &
   16.24 & 
   11.16 &
   17.75 &
    47.3 \\ 
\multicolumn{1}{c}{DRG \cite{Gao-ECCV-DRG}} &
  \multicolumn{1}{c}{R50-F} &
 24.53 &
   19.47 &
  \multicolumn{1}{c}{26.04} &
 51.0 \\ 
  \multicolumn{1}{c}{HOTR \cite{kim2021hotr}} &
  \multicolumn{1}{c}{R50} &
     25.10 &
  17.34 &
  \multicolumn{1}{c}{27.42} &
  55.2 \\
  \multicolumn{1}{c}{QPIC \cite{tamura_cvpr2021}} &
  \multicolumn{1}{c}{R50} &
     29.07 &
    21.85 &
  \multicolumn{1}{c}{31.23} &
   58.8  \\ 
   \multicolumn{1}{c}{$\textrm{CPC}$ \cite{park2022consistency}} &
  \multicolumn{1}{c}{R50} &
     29.63 &
   23.14 &
  \multicolumn{1}{c}{31.57} &
   63.1  \\ 
   \multicolumn{1}{c}{FGAHOI \cite{Ma2023FGAHOI}} &
  \multicolumn{1}{c}{Swin-t} &
   29.94 &
    22.24 &
  \multicolumn{1}{c}{32.24} &
    \multicolumn{1}{c}{60.5} \\
\multicolumn{1}{c}{RLIP \cite{yuan2022rlip}} &
  \multicolumn{1}{c}{R50} &
   32.84 &
    26.85 &
  \multicolumn{1}{c}{34.63} &
   61.9 \\ 
   \multicolumn{1}{c}{$\textrm{GENVLKT}$~\cite{liao2022gen}} &
  \multicolumn{1}{c}{R50} &
       33.75 &
   29.25 &
  \multicolumn{1}{c}{35.10} &

    \multicolumn{1}{c}{62.4} \\ 
\multicolumn{1}{c}{PViC \cite{pvic_Zhang_2023_ICCV}} &
  \multicolumn{1}{c}{R50} &
      34.69 
      & 32.14 
      & 35.45 
       & 62.8 
      \\ \multicolumn{1}{c}{$\textrm{HOICLIP}$ \cite{ning2023hoiclip}} &
  \multicolumn{1}{c}{R50} &
      34.69 
      & 31.12 
      & 35.74 
      & 63.5 
      \\
    \multicolumn{1}{c}{RLIPv2 \cite{Yuan_2023_ICCV}} &
   \multicolumn{1}{c}{R50} &
  35.38 &  
  29.61 &  
  37.10 & 65.9  \\ \multicolumn{1}{c}{LOGICHOI\cite{li2023neurallogic}} &
   \multicolumn{1}{c}{R50} &
  35.47 &  
  32.03 &  
  36.22 & 64.4 \\
  \multicolumn{1}{c}{RmLR\cite{Cao_2023_ICCV}} &
  \multicolumn{1}{c}{R50} &
     {36.93} 
     & 29.03 
     & {39.29} 
 & 63.7 
     \\ \multicolumn{1}{c}{ViPLO \citep{park2023viplo}} &
   \multicolumn{1}{c}{ViT-B} 
  & 37.22 
  & 35.45  
  & 37.75 & 62.2 \\
                \multicolumn{1}{l}{SCTC \cite{jiang2024exploringselfcrosstripletcorrelations}} &
  \multicolumn{1}{c}{R50} &
     37.92 
      & 34.78 
      & 38.86 
      & 67.1
      \\ 
  \multicolumn{1}{c}{ADA-CM \cite{ada_cm}} &
  \multicolumn{1}{c}{ViT-L} &
     38.40 
      & 37.52 
      & 38.66 
      &58.5 
      \\ 
             \multicolumn{1}{l}{BCOM \cite{bcom_Wang_2024_CVPR}} &
  \multicolumn{1}{c}{R50 } &
     39.34 
      & 39.90 
      & 39.17 
       & 65.8 
      \\ 
            \multicolumn{1}{c}{UniHOI \cite{unihoi}} &
  \multicolumn{1}{c}{R50} &
     40.06 
      & 39.91 
      & 40.11 
     & 65.5 
      \\ 
      \rowcolor{black!5}
\multicolumn{1}{c}{ContextHOI} &
  \multicolumn{1}{c}{R50} &
    {41.82} & \textbf{43.91}   & {41.19}  &
   \multicolumn{1}{c}{66.1} \\   \rowcolor{black!5}
  \multicolumn{1}{c}{ContextHOI} &
  \multicolumn{1}{c}{R101} &
  \textbf{42.09} &
  \underline{42.41} &
  \multicolumn{1}{c}{\textbf{41.99}} &
    \multicolumn{1}{c}{\textbf{67.3}} \\ \hline 
\end{tabular}}
\caption{\textbf{Performance comparison with state-of-the-art on benchmark HICO-DET.} R50 and R50-F refers to ResNet50 and ResNet50-FPN, Swin-t refers to Swin-tiny.}
  \label{tab:performance_hico}
\end{table}
\section{Experiments}
\subsection{Benchmarks and Metrics}
\label{sec:4.1}
\textbf{Traditional benchmarks.} We first conduct experiments on two widely-used HOI detection benchmark HICO-DET~\cite{18wacvHOI} and v-coco~\cite{gupta2015visual_vcoco}. HICO-DET comprises 80 object categories, 117 interaction categories, and 600 HOI triplet categories. HICO-DET contains 38,118 training images and 9,658 validation images. v-coco contains 5,400 train-val images and 4,946 validation images, consisting of 80 object categories, 29 verb categories, and 263 interaction triplet combinations. 

\noindent\textbf{Benchmark with ambiguous scenes.} In this section, we introduce a specialized benchmark designed to evaluate the robustness of models when handling HOI samples with unclear instance attributes, termed HICO-DET (ambiguous). We select a subset from the test set of HICO-DET~\cite{18wacvHOI} that includes images with unclear visual attributes.  Specifically, we engaged independent human volunteers to select images featuring unseen subjects, occluded subjects, blurred subjects, and instances too small to distinguish. A total of 659 images were selected, and together with their original annotations, they comprise the ambiguous benchmark.

\noindent\textbf{Metrics.} The mean Average Precision (mAP) is utilized for performance evaluation, following \cite{18wacvHOI}. For HICO-DET, mAP on both Default is reported. It has three different category settings, including $full$ for all 600 HOI categories, $rare$ for long-tail categories with less than 10 training samples, and the last \textrm{$non$-$rare$} categories. For v-coco, the role mAP in Scene 1 with 29 verb categories is reported.
\begin{table*}[]
\centering
\scalebox{0.80}{
\begin{tabular}{lccccccc} \hline
\multicolumn{1}{c}{\multirow{2}{*}{Method}} &
  \multicolumn{1}{c}{\multirow{2}{*}{Context}} &
  \multicolumn{3}{c}{HICO-DET (default)} & \multicolumn{3}{c}{HICO-DET (Ambiguous)}  \\ \multicolumn{1}{c}{} & \multicolumn{1}{c}{} &
 \multicolumn{1}{c}{\textit{Full}} &
  \multicolumn{1}{c}{\textit{Rare}} &
  \multicolumn{1}{c}{\textit{Non-rare}} & \multicolumn{1}{c}{\textit{Full}} &
  \multicolumn{1}{c}{\textit{Rare}} &
  \multicolumn{1}{c}{\textit{Non-rare}} \\
 \midrule
\multicolumn{5}{l}{Two-stage HOI Detectors}  \\ \midrule
UPT \cite{zhang2022upt}  & no context  & 31.66 & 25.94 & 33.36   & 16.53 (\textcolor{red}{-15.13}) & 10.70 (\textcolor{red}{-15.24}) & 18.27 (\textcolor{red}{-15.09}) \\
ADA-CM \cite{ada_cm}  & heuristic & 38.40 & 37.52 & 38.66   & 18.12 (\textcolor{red}{-20.28}) & 9.76 (\textcolor{red}{-27.76}) & 20.62 (\textcolor{red}{-18.94}) \\
\midrule
\multicolumn{5}{l}{One-stage HOI Detectors}  \\ \midrule
QPIC \citep{tamura_cvpr2021}  & no context & 29.07 & 21.85 & 31.23 & 9.25 (\textcolor{red}{-19.82}) & 6.18 (\textcolor{red}{-15.67}) & 9.61 (\textcolor{red}{-21.62})\\ 
ContextHOI  & ours & {41.82} & {43.91}   & {41.19} & \textbf{46.99} (\textcolor{green}{+5.17}) & \textbf{60.57} (\textcolor{green}{+16.66}) & \textbf{45.37} (\textcolor{green}{+4.18}) \\  \hline
\end{tabular}}
\caption{\textbf{Performance comparison with state-of-the-art on HICO-DET (default) and HICO-DET (ambiguous).} ContextHOI shows robustness on images with unclear instances.}
\label{tab:performance_hico_ambiguous}
\end{table*}
\subsection{Implementation Details}
\label{sec:4.2}
 We implement ContextHOI with transformer detectors introduced by DETR~\cite{detr}, and both ResNet50 and ResNet101 backbone. Our detector query dimension $N_q$ is 64 for both HICO-DET and v-coco. The encoders in the feature extractor have 6 layers, and the instance decoder, context extractor and context aggregator are all implemented by adopting a 3-layer transformer decoder. The transformer hidden dimension $C$ of all the components is 256. 
 
 We select CLIP ViT-L/14~\cite{clip} as our semantic teacher following~\cite{ada_cm}, several single linear layers with $\textrm{LayerNorm}$ is learned to match the size of transformer hidden dimension and CLIP visual linguistic dimensions. Our prediction heads follow the setting of~\cite{ning2023hoiclip}. The simple MLP predicting context region boxes is a 3-layer MLP. The similarity predictors in the semantic-guided context explorer are implemented by 3-Layer MLP, we initialize their weights with the CLIP text embedding of category prompts. The learnable parameter $\tau$ in $\mathcal{L}_{IC}$ is initialized to 0.5.

 ContextHOI is trained for 60 epochs with an AdamW optimizer with an initial learning rate of 1e-4 and a 10 times weight decay at 40 epochs. We train the model initialized with DETR~\cite{detr} params pre-trained on MS-COCO. The Hungarian matching process and cost coefficients follow the setting of \cite{tamura_cvpr2021}. We remain the coefficient setting of conventional HOI prediction losses in QPIC~\cite{tamura_cvpr2021}, for the spatial constraint losses, we set the loss coefficients $\lambda_{fc}, \lambda_{rc} $ and $ \lambda_{ic}$ to 4, 1 and 4, respectively. ContextHOI is trained on a single Tesla A100 GPU with batch size 16.

\begin{table*}[]
\centering
\begin{tabular}{ccc}
\begin{subtable}{.25\linewidth}
\centering
\subcaption{}
\scalebox{0.7}{
\begin{tabular}{ccc|ccc}
\midrule \multicolumn{1}{c}{Context} & \multicolumn{1}{c}{$\mathcal{L}_{SC}$} &  \multicolumn{1}{c|}{SCE} &
 \multicolumn{1}{c}{{Full}} &
  \multicolumn{1}{c}{{Rare}} &
  \multicolumn{1}{c}{{Non-rare} } \\ \midrule   \rowcolor{black!5}
   \multicolumn{1}{c}{} &  \multicolumn{1}{c}{} & \multicolumn{1}{c|}{} &
     34.69 
      & 31.12 
      & 35.75 
      \\  \rowcolor{white}
      \multicolumn{1}{c}{\ding{52}} &  \multicolumn{1}{c}{} & \multicolumn{1}{c|}{} &
     35.81 
      & 33.27 
      & 36.87 
      \\  \rowcolor{black!5}
  \multicolumn{1}{c}{\ding{52}} &   \multicolumn{1}{c}{\ding{52}} & & 
     36.75 
      & 35.47 
      & 37.12 
      \\   \rowcolor{white}
  \multicolumn{1}{c}{\ding{52}} &   \multicolumn{1}{c}{} & \ding{52}  &
     38.91 
      & 39.73 
      & 38.66 
      \\   \rowcolor{black!5}
  \multicolumn{1}{c}{\ding{52}} &   \multicolumn{1}{c}{\ding{52}} & \ding{52} & \textbf{41.82} & \textbf{43.91}   & {41.19} 
      \\\bottomrule
\end{tabular}}
\end{subtable}
\begin{subtable}{.25\linewidth}
\centering
\subcaption{}
\scalebox{0.75}{
\begin{tabular}{lccc}
\midrule                   \multicolumn{1}{c}{\multirow{1}{*}{$\mathcal{L}_{SC}$}}             & full  & rare & non-rare    \\ \midrule \rowcolor{black!5}
          None      &  38.91 
      & 39.73 
      & 38.66 
      \\  \rowcolor{white}
                             w/ $\mathcal{L}_{FC}$    &     40.93   &    41.56       &   40.74 \\ \rowcolor{black!5}
 w/ $\mathcal{L}_{RC}$ &  41.41   &    41.90       &    \textbf{41.26}        \\   \rowcolor{white}
 w/ $\mathcal{L}_{IC}$ &    40.59   &   40.77      & 40.54
                           \\   \rowcolor{black!5}
                           w/ $\mathcal{L}_{IC}$ + $\mathcal{W}_{d}$ &    41.46   &   42.84         & 41.05
                           \\ \rowcolor{white}
                           all   &   \textbf{41.82} & \textbf{43.91}   & {41.19}\\\hline 
                           \end{tabular}}
\end{subtable}
\begin{subtable}{.25\linewidth}
\centering
\subcaption{}
\scalebox{0.85}{
\begin{tabular}{lccc}
\midrule                   \multicolumn{1}{c}{\multirow{1}{*}{Prior}}      
                   & full  & rare & non-rare \\ \hline  \rowcolor{black!5}
        None   & 36.75   & 35.47  & 37.12  \\  \rowcolor{white}
    EVA-01       & 40.28   & 40.02  & 40.36            \\  \rowcolor{black!5}
                            CLIP-L/14   &   \textbf{41.82} & \textbf{43.91}   & {41.19}                      \\  \hline
                           \end{tabular}}

\end{subtable} 
\begin{subtable}{.25\linewidth}
\centering
\subcaption{}
\scalebox{0.76}{\begin{tabular}{cccc}
            \midrule
            \multicolumn{1}{c}{Aggregator} & full  & rare & non-rare \\ \hline       \rowcolor{black!5}
               w/ $\mathbf{Z}_{c}$ &  39.92   & 40.18  & 39.84   \\ \rowcolor{white}
                     w/ $\mathbf{Z}_{c}$ + $\mathbf{Z}_{v}$     &   \textbf{41.82} & \textbf{43.91}   & \textbf{41.19} \\
         \hline
        \end{tabular}}
\end{subtable} 
\end{tabular}
\caption{\textbf{Ablations} of (a) Proposed context-aware supervision, SCE refers to Semantic-guided explorers, (b) components of spatially contrastive constraints, (c) utilizing prior knowledge from different VLMs, and (d) the implementation of the context aggregator. All experiments are conducted with ResNet50 backbone.}
  \label{table:ablation_others}
\end{table*}

\subsection{Comparison with State-of-the-Art Methods.}
\label{sec:4.3}
\noindent\textbf{Regular HOI prediction.} We provide performance comparisons with the state-of-the-art on regular settings of HICO-DET and v-coco. As detailed in Table~\ref{tab:performance_hico}, ContextHOI with a ResNet50 backbone has 41.82 $full$ mAP and a higher 43.91 mAP on the $rare$ split, achieves state-of-the-art. We hypothesize that the spatial context information helps to discover the long-tail HOIs. ContextHOI with a larger ResNet101 backbone has a higher 42.09 $full$ mAP. As for v-coco, ContextHOI achieves 66.1 mAP on AP role Scene 1 with ResNet50 and 67.3 mAP with ResNet101. Both performance outperforms existing HOI detectors under the same backbones.

 \begin{figure}[t]
  \centering
   \includegraphics[width=1.0\linewidth]{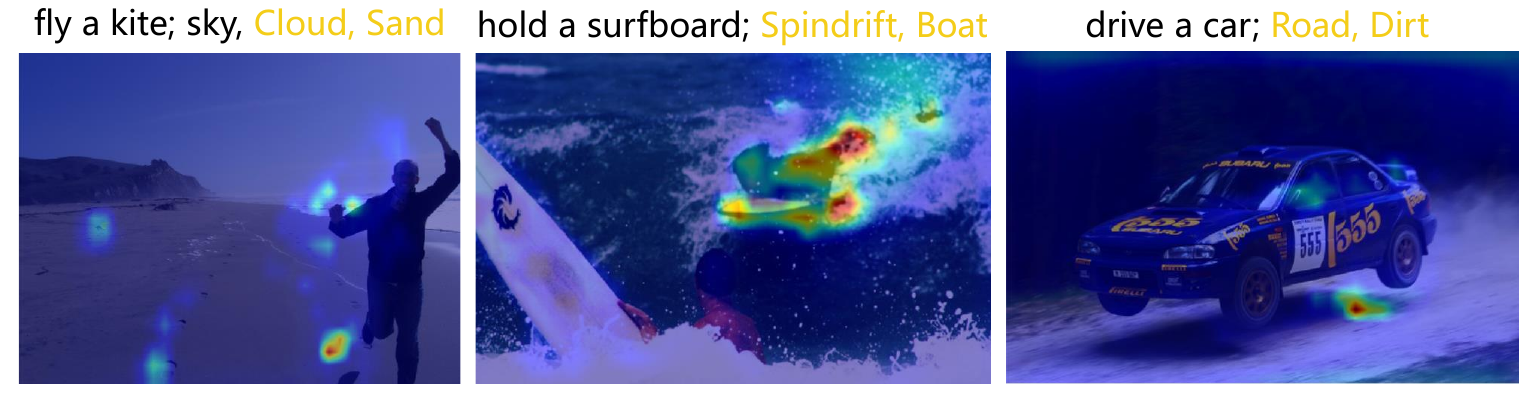}
   \caption{\textbf{Visualizations of the visual feature captured by ContextHOI on images in HICO-DET (ambiguous).} We mask the predicted instance boxes and let GPT-4V~\cite{gpt4_openai} describe the left images; the words describing contexts in GPT captions are selected and shown as the yellow light texts.}
   \label{fig:figure_intro_rebuttal}
\end{figure} 
\noindent\textbf{Ambiguous benchmarks.} We evaluate ContextHOI and several fundamental one-stage and two-stage detectors on HICO-DET (ambiguous). As demonstrated in Table~\ref{tab:performance_hico_ambiguous}, traditional one-stage and two-stage HOI detectors based on detection transformers show vulnerability in unclear and ambiguous scenes. The two-stage methods UPT~\cite{zhang2022upt} and ADA-CM~\cite{ada_cm}, which depend on pre-trained detection backbones, experience more performance decline than one-stage detectors, though they learn background information heuristically.

By integrating these techniques with our spatial context learning, our model further improves performance, showing a 5.17 increase in overall mAP and a 16.66 increase in rare mAP for ambiguous scenarios. This highlights the capability of our model to recognize interactions even with limited visual cues effectively. As ContextHOI also has a competitive performance on the regular HICO-DET test set, it indicates the spatial context learning do not excessively introduce irrelevant noise to the instance detection branch.
\subsection{Effectiveness of Spatial Context Learning}
\label{sec:4.4}
\noindent\textbf{Ablations of proposed modules.}
First, we provide performance ablation about our proposed spatial and semantic supervision in Table 3 (a). Directly equipping the base model~\cite{ning2023hoiclip} with an additional context branch can enhance the performance. The spatially contrastive constraints and the semantic-guided context exploration significantly further enhance the performance. Each type of supervision complements the other, with their combination yielding the highest mAP.

\noindent\textbf{Visualizations of spatial context.}
We further validate the effectiveness of spatial context learning with visualizations, as illustrated in Figure~\ref{fig:ablation_heatmap}, First, in (a), we demonstrate the captured features of the last layer of the instance decoder, the context aggregator and context extractor, presented from left to right.  Evidently, each component consistently focuses on the intended feature regions, aligning well with their respective design objectives.

In (b), we evaluate the effectiveness of our designed supervision by comparing the heatmaps of the context extractor under different supervision settings. From left to right, the first heatmap shows that when the context extractor is guided solely by semantic supervision, it primarily focuses on the instance regions and duplicates the function of the instance decoder. With only spatial constraints, the context extractor focuses on regions close to the margin, which introduces unexpected noise. However, when both types of supervision are applied, the context extractor captures informative context regions, which are well-shaped.

\noindent\textbf{Interpretability of spatial contexts.}
In the final block of Figure~\ref{fig:ablation_heatmap} (b), we present captions generated by GPT-4V~\cite{gpt4_openai} that describe the spatial contexts depicted in the image. GPT-4V identifies the surrounding environment as either a parking lot or a fairground. These labels align with the context map shown in (a) and intuitively support the prediction of $parking$. This alignment highlights the implicit semantic meanings of the captured context regions.

\subsection{Ablations}
\label{sec:4.5}
In this part, we provide detailed ablations on the model design and parameter efficiency of ContextHOI.

\noindent\textbf{Spatially contrastive constraints.} Table 3 (b) provides the ablation on the different settings of spatially contrastive constraints. All these constraint losses boost ContextHOI, while the naive $\mathcal{L}_{IC}$ only provides light enhancement. With our proposed dynamic distance weight, $\mathcal{L}_{IC}$ provides the highest enhancement. Combining all of these constraints, we obtain the best performance.

\noindent\textbf{Different VLMs as knowledge teacher.} Table 3 (c) compares different pre-trained VLMs utilized as our prior knowledge teacher. We try to train ContextHOI with no prior knowledge, with the visual-linguistic knowledge of EVA-CLIP-01~\cite{EVA, EVA-CLIP} and with CLIP ViT-L/14~\cite{clip}. The OpenAI CLIP is the best VLM fitting our model.

\noindent\textbf{Implementation of context aggregator.} We evaluate different implementations of mechanisms integrating the instance and context featured in Table 3 (d). The term add denotes a gated addition approach between $\mathbf{Z}_{ins}$ and $\mathbf{Z}_{c}$, enhancing feature integration by applying a learned gate to control the contribution from each feature set. The cross-attn term indicates shared cross-attention, as detailed in Section 3.4. In the last row, we further enhance the feature aggregation by incorporating visual features, $\mathbf{Z}_{v}$, obtained from a pre-trained VLM visual encoder into the shared cross-attention mechanism. The auxiliary semantic guidance provided by $\mathbf{Z}_{v}$ also contributes to the robustness of the feature aggregation process.

\noindent\textbf{Model efficiency.} 
We evaluate model efficiency by conducting inference on the first 100 HICO-DET~\cite{18wacvHOI} test images with resolution 800$\times$1333, batch size 1, on a single A800 GPU. ContextHOI maintains an acceptable 176.43 GFLOPs and 13.37 FPS, compared with HOICLIP~\cite{ning2023hoiclip} with 159.93 GFLOPs and 16.47 FPS. Our context-aware supervision, as training losses, contributes a significant performance increase with light additional parameters, and it is not required during inference. This design achieves a balance between computational cost and performance gains.

\noindent\textbf{More visualizations.} 
We show additional visualizations of the fused feature map for interaction prediction of ContextHOI in Figure~\ref{fig:figure_intro_rebuttal}. While recognizing images with tiny objects, unclear subjects and occluded subjects, ContextHOI tends to focus on spatial backgrounds containing informative environments. For instance, the sand and cloud for $fly$ a $kite$, wave for $hold$ a $surfboard$, and the dirt road for $dirve$ a $car$.


\section{Conlusion \& Discussion}
We introduce a novel spatial context learning paradigm tailored for transformer-based HOI detectors, featuring a dual-branch and fusion architecture. This framework is augmented with novel spatially contrastive constraints and a semantic-guided context explorer. The proposed components combine our network ContextHOI, achieving a new state-of-the-art performance on HICO-DET with ResNet50. ContextHOI demonstrates significant robustness in HOI scenarios with unclear and occluded instance clues. In the future, we plan to explore more fine-grained context-learning approaches to better mitigate background noise for robust HOI detection.
\clearpage
\bibliography{aaai25}
\end{document}